\documentclass{article}
\usepackage{ijcai18}
\usepackage{graphicx}  %Required
\usepackage{subfigure}
\usepackage{float}
\usepackage{amsmath}
\usepackage{amsthm}
\usepackage{bbding}
\usepackage{hyperref}
\usepackage{footnote}
\usepackage[T1]{fontenc}
\usepackage{aecompl}
\usepackage{amssymb}
\usepackage{bbding}
\usepackage{pifont}
\newcommand{\cmark}{\ding{52}}
\newcommand{\xmark}{\ding{56}}

\usepackage{times}
\usepackage{soul}
\usepackage[utf8]{inputenc}
\usepackage[small]{caption}

\title{IncepText: A New Inception-Text Module with Deformable PSROI Pooling for Multi-Oriented Scene Text Detection}

\author{}

\iftrue
\author{
	Qiangpeng Yang, 
	Mengli Cheng, 
	Wenmeng Zhou,
	Yan Chen,
	Minghui Qiu,
	Wei Lin, 
	Wei Chu
	\\ 
	Alibaba Group \\
	\{qiangpeng.yqp, mengli.cml, wenmeng.zwm, chenyan.cy, minghui.qmh, weilin.lw, weichu.cw\}@alibaba-inc.com
}
\fi

\begin{document}
	
\maketitle

\begin{abstract}
	Incidental scene text detection, especially for multi-oriented text regions, is one of the most challenging tasks in many computer vision applications.
	Different from the common object detection task, scene text often suffers from a large variance of aspect ratio, scale, and orientation. To solve this problem, we propose a novel end-to-end scene text detector IncepText from an instance-aware segmentation perspective. 
	We design a novel Inception-Text module and introduce deformable PSROI pooling to deal with multi-oriented text detection. 
	Extensive experiments on ICDAR2015, RCTW-17, and MSRA-TD500 datasets demonstrate our method's superiority in terms of both effectiveness and efficiency.
	Our proposed method achieves 1st place result on ICDAR2015 challenge and the state-of-the-art performance on other datasets. Moreover, we have released our implementation as an OCR product which is available for public access.
	%\footnote{The link is anonymized for blind review, which will be made available afterward.}.
	\footnote{\url{https://market.aliyun.com/products/57124001/cmapi020020.html}}
\end{abstract}

\section{Introduction}
%% brief introduction
Scene text detection is one of the most challenging tasks in many computer vision applications such as multilingual translation, image retrieval, and automatic driving. The first challenge is scene text contains various kinds of images, such as street views, posters, menus, indoor scenes, \emph{etc}. Furthermore, the scene text has large variations in both foreground texts and background objects, and also with various lighting, burring, and orientation.

%% method introduction
In the past years, there have been many outstanding approaches focus on scene text detection. The key point of text detection is to design features to distinguish text and non-text regions. Most of the traditional methods such as MSER \cite{neumann2010method} and FASText \cite{busta2015fastext} use manually designed text features. %design text features manually. 
These methods are not robust enough to handle complex scene text. Recently, Convolutional Neural Network (CNN) %~\cite{lecun-bengio-95a,lecun-98b} 
based methods achieve the state-of-the-art results in text detection and recognition \cite{he2016text,tian2016detecting,zhou2017east,he2017deep}. CNN based models have a powerful capability of feature representation, and deeper CNN models are able to extract higher level or abstract features.

In the literature, there are mainly two types of approaches for scene text detection, namely indirect and direct regressions. Indirect regression methods predict the offsets from some box proposals, such as CTPN~\cite{tian2016detecting} and RRPN \cite{ma2017arbitrary}. These methods are based on Faster-RCNN \cite{ren2015faster} framework. Recently, direct regression methods have achieved high performance for scene text detection, e.g. East \cite{zhou2017east} and DDR \cite{he2017deep}. Direct regression usually performs boundary regression by predicting the offsets from a given point.

In this paper, we solve this problem from an instance-aware segmentation perspective that mainly draws on the experience of FCIS \cite{li2016fully}. Different from common object detection, scene text often suffers from a large variance of scale, aspect ratio, and orientation. Therefore, we design a novel Inception-Text module to deal with these challenges. This module is inspired by Inception module \cite{szegedy2015going} in GoogLeNet, we choose multi branches of different convolution kernels to deal with the text of different aspect ratios and scales. At the end of each branch, we add a deformable convolution layer to adapt multi orientations. Another improvement is that we replace the PSROI pooling in FCIS with deformable PSROI pooling \cite{dai2017deformable}. According to our experiments, deformable PSROI pooling has better performance in the classification task.

Our main contributions can be summarized as follows:
\begin{itemize}
	\item We propose a new Inception-Text module for multi-oriented scene text detection. According to our experiments, this module shows a significant increase in accuracy with little computation cost.
	\item We propose to use deformable PSROI pooling module to deal with multi-oriented text. The qualitative study of learned offset parts in deformable PSROI pooling and quantitive evaluations show its efficiency to handle arbitrary oriented scene text.
	\item We evaluate our proposed method on three public datasets ICDAR2015, RCTW-17 and MSRA-TD500, and show that our proposed method achieves the state-of-the-art performance on several benchmarks without using any extra data.
	\item Our proposed method has been implemented as an API service in our OCR product, which is available in public.
\end{itemize}

The rest of this paper is organized as follows: we first give a brief overview of scene text detection and mainly focus on multi-oriented scene text detection. Then we describe our proposed method in detail and present experimental results on three public benchmarks. We conclude this paper and discuss future work at the end.

\section{Related Work}

Scene text detection has been extensively studied in the last decades. Most of the previous work focused on horizontal text detection, while more recent research studies have concentrated on multi-oriented scene text detection. Below we briefly introduce the related studies. %In this section, we introduce some awesome works on multi-oriented scene text detection.

\noindent \textbf{HMP.} HMP \cite{yao2016scene} is inspired by Holistically-Nested Edge Detection (HED) \cite{xie2015holistically}. It simultaneously predicts the probability of text regions, characters and the relationship among adjacent characters with a unified framework. Therefore, two kinds of label maps are needed: the label map of text line and the label map of characters. Graph partition algorithm is used to determine the retained and eliminated linkings, which is not robust enough for scene text detection.

\noindent \textbf{SegLink.} SegLink \cite{shi2017detecting} introduced a novel text detection framework which decomposes the text into two locally detectable elements, segments and links. A segment is an oriented box of a text line, while a link indicates the two adjacent segments belong to the same text line or not. The segments and links are detected at multiple scales by a fully convolutional network. However, a post-process step of combining segments is also needed in SegLink.

\noindent \textbf{RRPN.} RRPN \cite{ma2017arbitrary} is modified from Faster-RCNN \cite{ren2015faster} for multi-oriented scene text detection. The main difference between RRPN and Faster-RCNN is that anchors with six different orientations are generated at each position of the feature map. The angle information is a regression target in regression task to get more accurate boxes. 

\noindent \textbf{EAST.} EAST \cite{zhou2017east} proposed an efficient scene text detector which uses a single fully convolutional network. The network has two branches: a segmentation task predicts the text score map and a regression task which directly predicts the final box for each point in the text region. According to our experiments, this framework is not suitable for long text line, maybe a line grouping method is needed in post-processing. 

\noindent \textbf{DDR.} Deep Direct Regression (DDR) \cite{he2017deep} is very similar to EAST. They use a fully convolutional network to directly predict the final quadrilateral from a given point. In testing, a multi-scale sliding window strategy is used, which is very time-consuming. The main limitation is the same as EAST.

In a nutshell, different from previous models, our method is an end-to-end trainable neural network from an instance-aware segmentation perspective. We design an new Inception-Text module for multi-oriented text detection. To handle arbitrary oriented text, we replace standard PSROI pooling with deformable PSROI pooling and demonstrate its efficiency. Below we present our method in detail.

\section{The Proposed Method}
\subsection{Overview}
\begin{figure*}[th]
	\centering
	\includegraphics[width=0.82\textwidth]{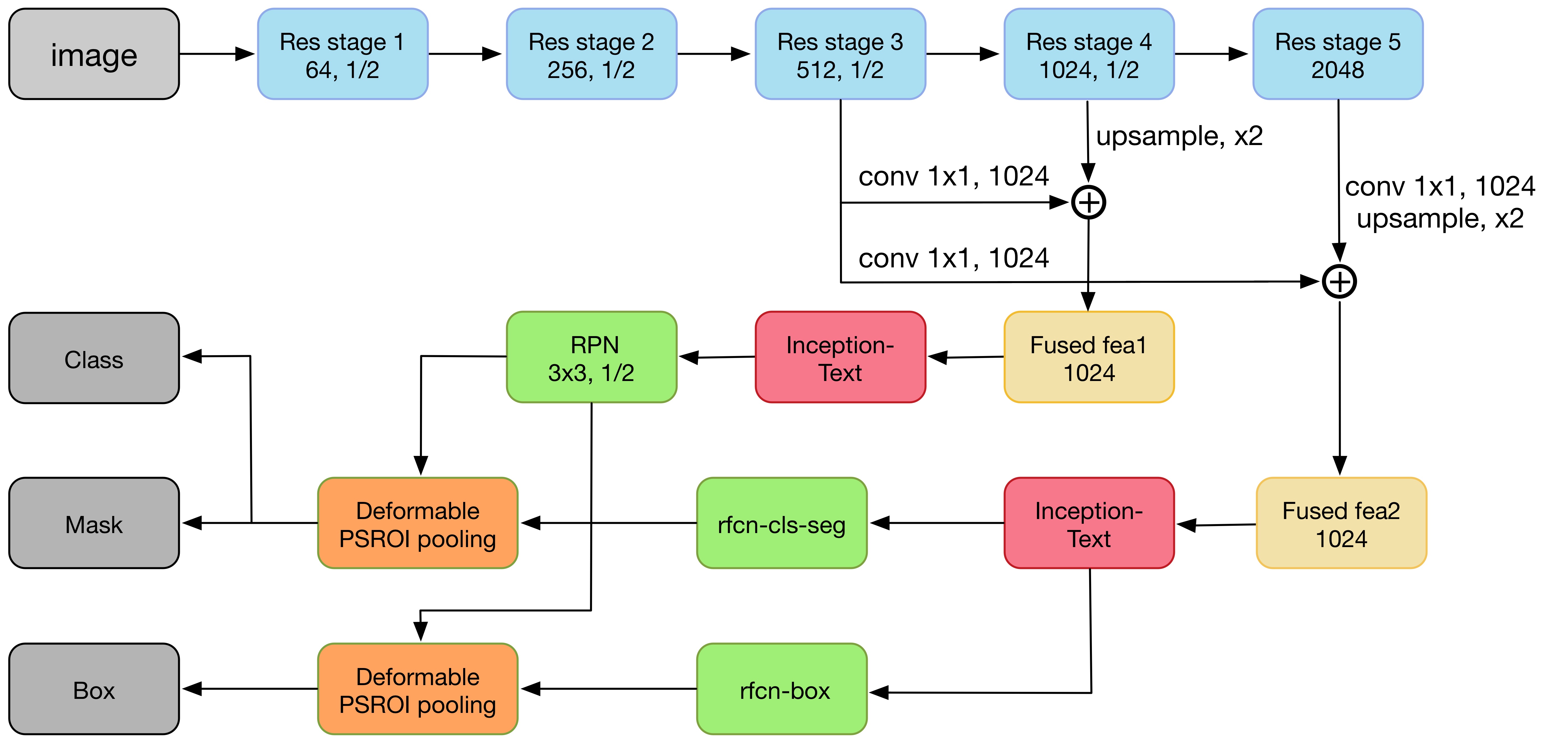}
	\caption{An overview of IncepText architecture. The basic feature extraction module in this figure is ResNet-50. Inception-Text module is appended after feature fusion, and the original PSROI pooling is replaced by deformable PSROI pooling.}
	\label{fig:network}
\end{figure*}

Our proposed method is based on FCIS \cite{li2016fully} framework, which is originally proposed for instance-aware segmentation. We design a novel Inception-Text module and use deformable PSROI pooling to extend this framework for scene text detection.  Figure \ref{fig:network} shows an overview of our model architecture.

In Figure \ref{fig:network}, the basic feature extraction module is ResNet-50 \cite{he2016deep}. For scene text detection, finer feature information is very important especially for segmentation task, the final downsampling in res stage 5 may lose some useful information. Therefore we exploit hole algorithm \cite{long2015fully} in res stage 5 to maintain to field of view. The stride=2 operations in the stage are changed to stride=1, and all convolutional filters on the stage use hole algorithm instead to compensate the reduced stride.

To predict accurate location of small text regions, low-level features also need to be taken into consideration. 
As illustrated in Figure~\ref{fig:network}, layer \emph{res4f} and layer \emph{res5c} are upsampled by a factor 2 and added with layer \emph{res3d}. Then these two fused feature maps are followed by Inception-Text module which is designed for scene text detection. We replaced PSROI pooling layer in FCIS with deformable PSROI pooling, because standard PSROI pooling can only handle horizontal text while scene text always has arbitrary orientations.
Similar to FCIS, we obtain text boxes with masks and classification scores as in Figure 1, and then we apply NMS on the boxes based on their scores. For each unsuppressed boxes, we find its similar boxes which are defined as the suppressed boxes that overlap with the unsuppressed box by $IoU>=0.5$. The prediction masks of the unsuppressed boxes and its similar boxes are merged by weighted averaging, pixel-by-pixel, using the classification scores as their averaging weights. And then a simple minimal quadrilateral algorithm is used to generate the oriented boxes.

\subsection{Inception-Text}

\begin{figure}[th]
	\centering
	\includegraphics[width=0.43\textwidth]{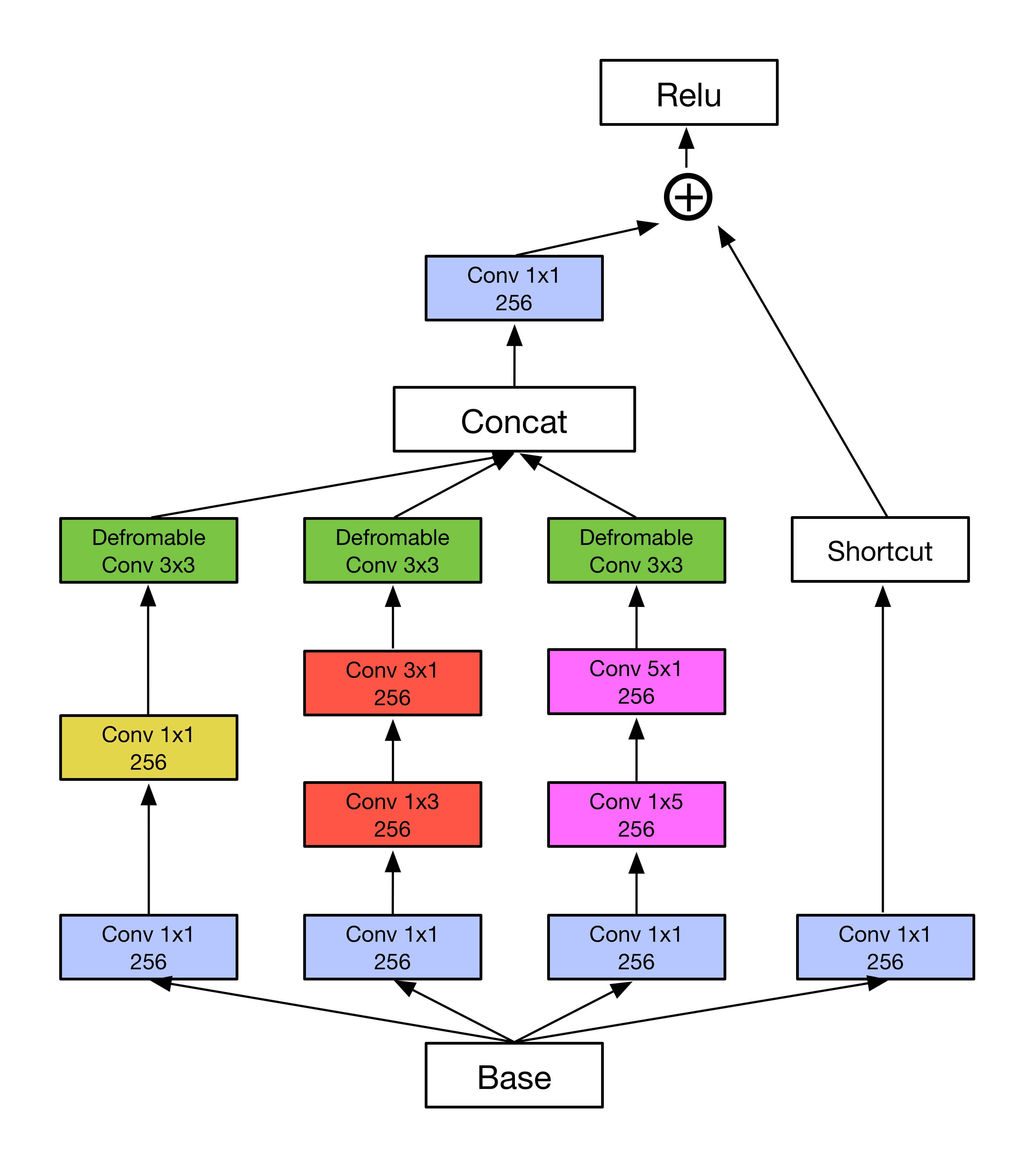}
	\caption{Inception-Text module.}
	\label{fig:inception_text}
\end{figure}

Our proposed Inception-Text module mainly has two parts, which is shown in Figure \ref{fig:inception_text}. The first part has three branches which is very similar to Inception module in GoogLeNet. Firstly, we add a $1\times1$ conv-layer to decrease the number of channels in the original feature map. To deal with the different scales of text, three different convolution kernels: $1\times1$, $3\times3$ and $5\times5$ are selected. Different scales of text are activated in different branches. 
%Considering that scene text suffers a large variance of aspect ratio, 
We further factorize the $n\times n$ convolution into two convolutions, which is a $1\times n$ convolution followed by a $n\times1$ convolution. According to \cite{szegedy2016rethinking}, these two structures have same receptive fields, while the factorization has a lower computational cost.

\begin{figure}[h]
	\centering
	\subfigure[standard Convolution]{\includegraphics[width=0.4\textwidth]{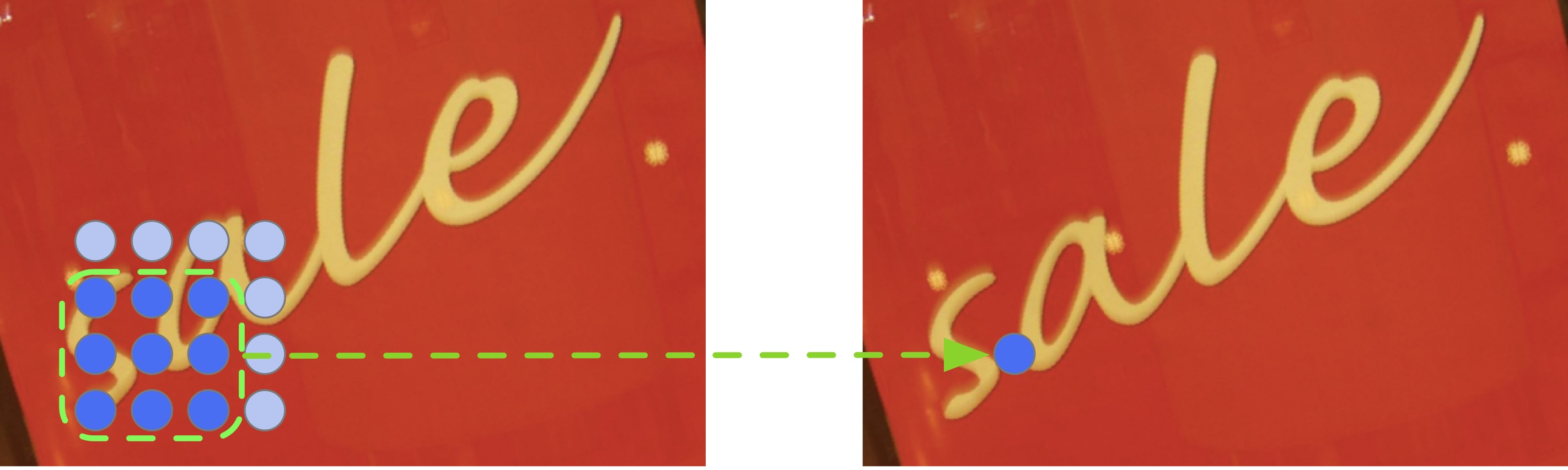}}
	\subfigure[Deformable Convolution]{\includegraphics[width=0.4\textwidth]{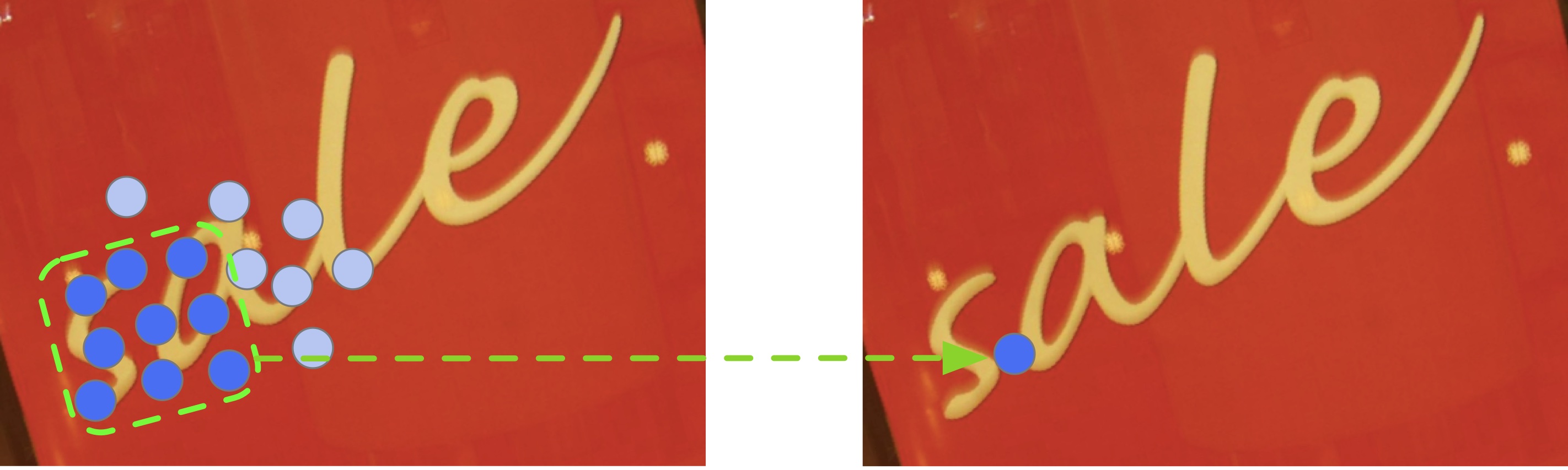}}
	\caption{Comparison between standard convolution and deformable convolution. The receptive field in standard convolution (a) is fixed while deformable convolution (b) has adaptive receptive field.}
	\label{fig:deform_conv}
\end{figure}

Comparing to standard Inception module, another important difference is that we add a deformable convolution layer at the end of each branch of the first part. Deformable convolution layer is firstly introduced in \cite{dai2017deformable}, where the spatial sampling locations are augmented with additional offsets learned from data. In scene text detection, arbitrary text orientation is one of the most challenging problems, deformable convolution allows free form deformation of sampling grid instead of regular sampling grid in standard convolution. This deformation is conditioned over the input features, thus the receptive field is adjusted when the input text is rotated.  To illustrate this, we compare standard convolution and deformable convolution in Figure \ref{fig:deform_conv}. Clearly, the standard convolution layer can only handle horizontal text regions, while the deformable convolution layer is able to use an adaptive receptive field to capture regions with different orientations. 
More quantitive results are illustrated in Table~\ref{table:Impact}.
%Hence deformable convolution layer can help to improve accuracy.

Furthermore, similar to Inception-ResNet V2 \cite{szegedy2017inception}, we also apply the shortcut design followed by a $1\times1$ conv-layer.

\subsection{Deformable PSROI Pooling}

PSROI pooling \cite{dai2016r} is a variant of regular ROI pooling, which operates on position-sensitive score maps with no weighted layers following. The position-sensitive property encodes useful spatial information for classification and object location. 

However, for multi-oriented text detection task, PSROI pooling can only deal with axis-aligned proposals. Hence we use deformable PSROI pooling \cite{dai2017deformable} to add offsets to the spatial binning positions in PSROI pooling. These offsets are learned purely from data. The deformable PSROI pooling is defined as:

\begin{equation}
r_{c}(i, j) = \sum_{(x, y) \in bin(i, j)} \frac{\hat{z}_{i, j}(x+x_{0}+\Delta x, y+y_{0}+
	\Delta y)}{n},
\end{equation}
where $r_{c}(i, j)$ is the pooled response in the $(i, j)$-th bin, $\hat{z}_{(i, j)}$ is the transformed feature map, $(x_{0}, y_{0})$ is the top-left corner of an ROI, $n$ is the number of pixels in the bin. $\Delta x$ and $\Delta y$ are learned from a fc layer.

\begin{figure}[H]
	\centering
	\subfigure{\includegraphics[width=0.21\textwidth]{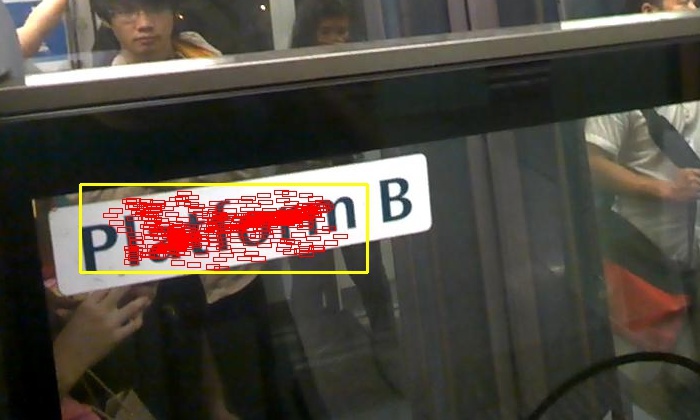}}
	\subfigure{\includegraphics[width=0.21\textwidth]{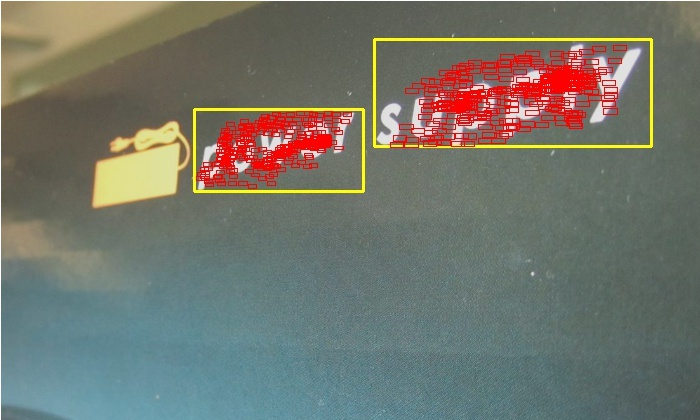}}
	\caption{Visualization of learned offsetted parts in deformable PSROI pooling. We have $21 \times 21$ bins (red) for the each input ROI (yellow). Deformable PSROI pooling tend to learn the context surrounding the text.}
	\label{fig:deform_psroi}
\end{figure}

Deformable PSROI Pooling is proposed for non-rigid object detection, and we apply it in multi-oriented scene text detection. In Figure \ref{fig:deform_psroi}, we take a brief visualization of how the parts are offset to cover the text with arbitrary orientation. More quantitative analyses are shown in Table \ref{table:Impact}.   

\subsection{Ground Truth and Loss Function}

\begin{figure}[H]
	\centering
	\includegraphics[width=0.35\textwidth]{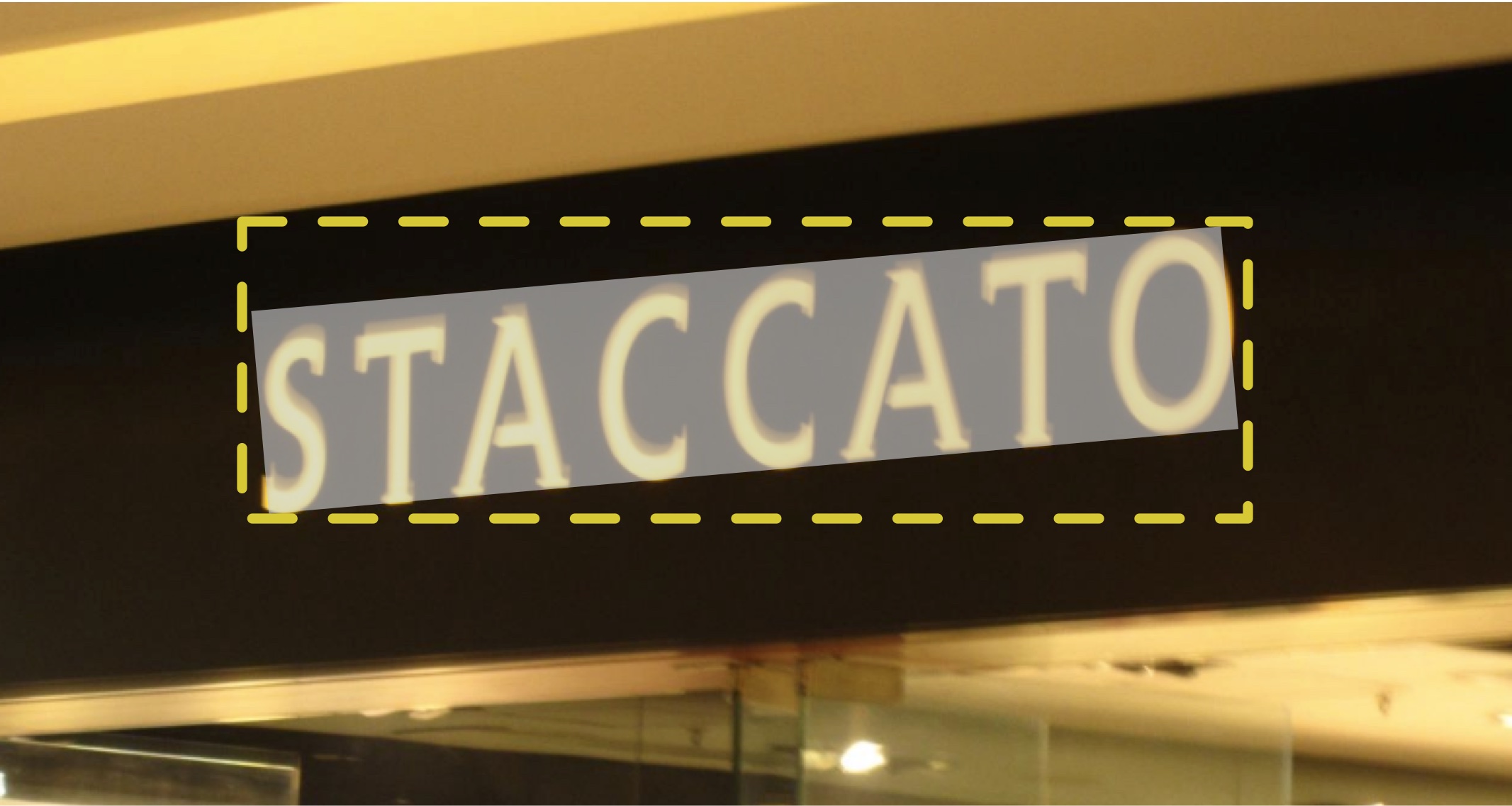}
	\caption{Ground Truth. The target of regression task is colored in yellow dashed lines, and the mask target is filled with gray quadrilateral.}
	\label{fig:ground_truth}
\end{figure}

The ground truth of text instance is exemplified in Figure \ref{fig:ground_truth}. Different from general instance-aware segmentation task, we do not have the pixel-wise label of text and non-text. Instead, the pixels in the quadrilateral are all positive, while the left pixels are negative.

The loss function is similar to FCIS \cite{li2016fully}, which can be formulated as:
\begin{equation}\label{equ:loss}
\begin{aligned}
L = L_{rcls} + L_{rbox} + L_{cls} + L_{box} + \lambda_{m}L_{mask}
%& L = L_{rpn} + L_{rcnn} + \lambda_{s}L_{s} \\
%& L_{rpn} = L_{rcls} + L_{rbox} \\
%& L_{rcnn} = L_{cls} + L_{box} \\
\end{aligned}
\end{equation}
where $L_{rcls}$ and $L_{rbox}$ are classification and regression loss in RPN stage, while $L_{cls}$ and $L_{box}$ are in RCNN stage. $L_{mask}$ is cross-entropy loss for segmentation task, where $\lambda_{m}$ in our experiments is set to $2$.

\section{Experiments}
\subsection{Benchmark Datasets}
We evaluated our method on three public benchmark datasets. These datasets all have scene text with arbitrary orientations.

%\subsubsection{ICDAR2015 Incidental Scene Text.}
\noindent \textbf{ICDAR2015.} This dataset was used in challenge 4 of ICDAR2015 Robust Reading competition \cite{karatzas2015icdar}. It contains 1000 images for training while 500 images for testing. These images were collected by Google Glass, which suffers from motion blur and low resolution. The bounding boxes of text have multi-orientations and they are specified by the coordinates of their four corners in a clock-wise manner.

%\subsubsection{RCTW-17.}
\noindent \textbf{RCTW-17.} This is a competition on reading Chinese Text in images, which contains a various kinds of images, including street views, posters, menus, indoor scenes and screenshots. Most of the images are taken by phone cameras. This dataset contains about 8000 training images and 4000 test images. Annotations of RCTW-17 are similar to ICDAR2015.

%\subsubsection{MSRA-TD500.}
\noindent \textbf{MSRA-TD500.} This was collected from indoor and outdoor scenes using a pocket camera~\cite{yao2012detecting}. This dataset contains 300 training images and 200 testing images. Different from ICDAR2015, the basic unit in this dataset is text line rather than word and the text line may be in different languages, Chinese, English, or a mixture of both.

\subsection{Experimental Setup}

Our proposed network was trained end-to-end using ADAM optimizer \cite{kingma2014adam}. We adopted the multi-step strategy to update learning rate, and the initial learning rate is $10^{-3}$. Each image is randomly cropped and scaled to have short edge of $\{640, 800, 960, 1120\}$. The anchor scales are $\{2, 4, 8, 16\}$, and ratios are $\{0.2, 0.5, 2, 5\}$. 
And we also applied online hard example mining (OHEM) for balancing the positive and negative samples.

\subsection{Experimental Results}
\subsubsection{Impact of Inception-Text and Deformable PSROI pooling.}

We conducted several experiments to evaluate the effectiveness of our design. These experiments mainly focus on evaluating two important modules in our model: Inception-Text and deformable PSROI pooling.  Table \ref{table:Impact} summarizes the results of our models with different settings on ICDAR 2015. 

\begin{figure}[H]
	\centering
	\subfigure[]{\includegraphics[width=0.15\textwidth]{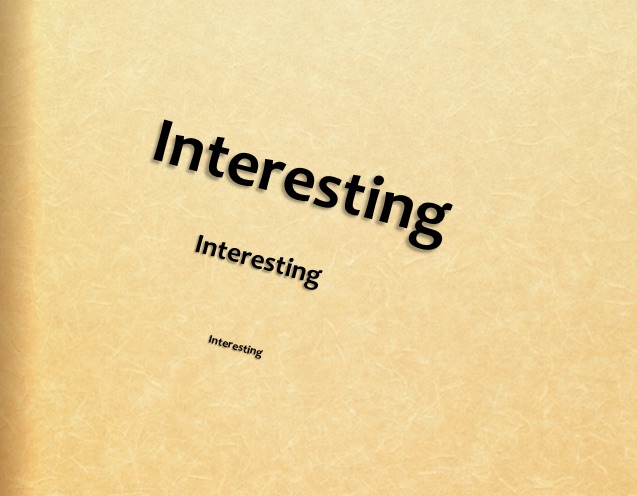}}
	\\
	\subfigure[]{\includegraphics[width=0.15\textwidth]{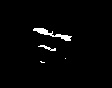}}
	\subfigure[]{\includegraphics[width=0.15\textwidth]{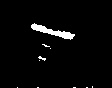}}
	\subfigure[]{\includegraphics[width=0.15\textwidth]{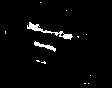}}
	\subfigure[]{\includegraphics[width=0.15\textwidth]{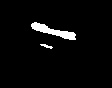}}
	\subfigure[]{\includegraphics[width=0.15\textwidth]{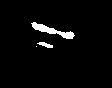}}
	\subfigure[]{\includegraphics[width=0.15\textwidth]{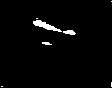}}
	\subfigure[]{\includegraphics[width=0.15\textwidth]{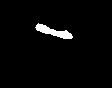}}
	\subfigure[]{\includegraphics[width=0.15\textwidth]{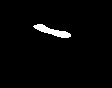}}
	\subfigure[]{\includegraphics[width=0.15\textwidth]{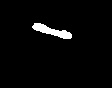}}
	\caption{Feature maps of each branch in Inception-Text module. (a) Input image. (b)(c)(d) Feature maps of the left branch. (e)(f)(g) Feature maps of the middle branch. (h)(i)(j) Feature maps of the right branch.}
	\label{fig:feamap}
\end{figure}

\paragraph{Inception-Text.} We designed this module to handle text with multiple scales, aspect ratios and orientations. To evaluate this module, we set the input image with text of three different scales and visualize the feature maps at the end of each branch. An interesting phenomenon is exemplified in Figure \ref{fig:feamap}. The left branch (kernel size = 1) in Figure \ref{fig:inception_text} is activated with three scales, some channels of the middle branch (kernel size = 3) are not activated with the smallest text, and some channels of the right branch (kernel size = 5) are only activated with the largest text. 

\begin{figure}[th]
	\centering
	\subfigure[]{\includegraphics[width=0.16\textwidth]{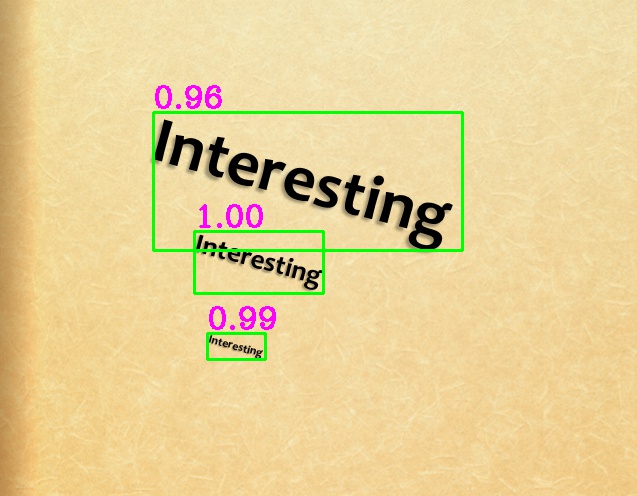}}
	\subfigure[]{\includegraphics[width=0.16\textwidth]{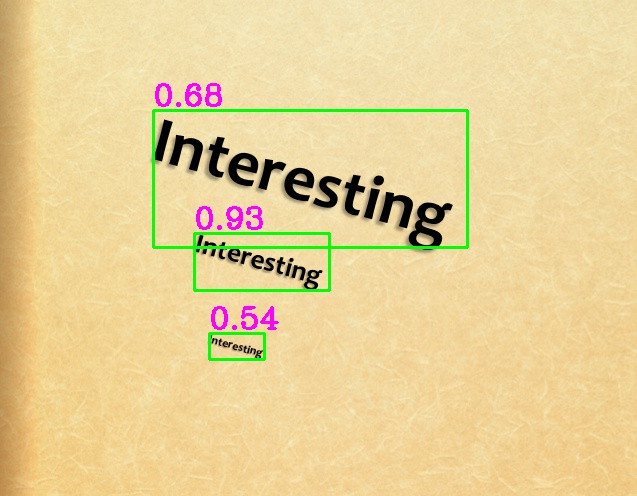}}
	\subfigure[]{\includegraphics[width=0.16\textwidth]{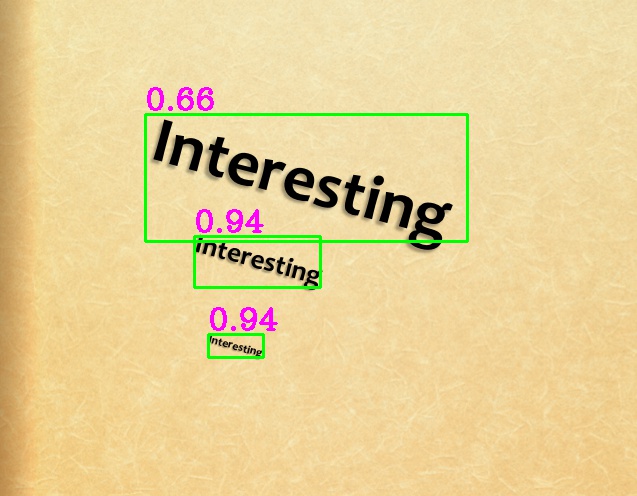}}
	\subfigure[]{\includegraphics[width=0.16\textwidth]{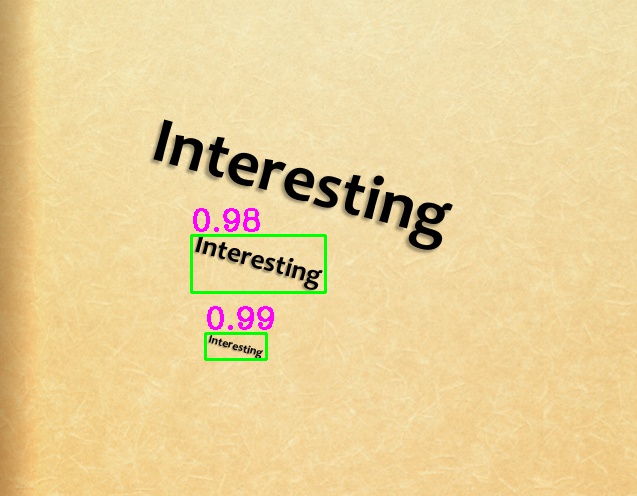}}
	\caption{Influence of different branch. (a) Result with three branches. (b) Result without the left branch. (b) Result without the middle branch. (c) Result without the right branch.}
	\label{fig:influence}
\end{figure}

We also conducted another experiment in Figure \ref{fig:influence}. We found that, if we used all three branches in testing, all words will be detected with high confidence. When we remove the left branch, the scores of three words are decreased simultaneously and the smallest word decreased farthest. If we remove the middle branch, the influence of the smallest word is reduced. When we remove the right branch, the biggest word is missed, while the other two words have little influence. %The experiment result is shown in 

%phenomenons
These two experiments demonstrate that different scales of text are activated in different branches. The branch with large kernel size has more influence on large text, while the small text is mainly influenced by branch of small kernel size.

In addition, we compared the origin Inception module and our Inception-Text module, our design has about $0.017$ improvement in recall (0.803 vs. 0.786) while the precision is almost the same. Comparing to the origin design (without Inception or Inception-Text), both recall (0.803 vs. 0.775) and precision (0.891 vs. 0.873) have large improvements. The final F-measure has over $0.02$ improvement.

\begin{table}[h]
	\centering
	\fontsize{8}{8}\selectfont
	\setlength{\tabcolsep}{4pt}
	\renewcommand{\arraystretch}{2.0}
	\newcommand{\tabincell}[2]{\begin{tabular}{@{}#1@{}}#2\end{tabular}}  
	\begin{tabular}{c|c|c|c|c|c}
		\hline
		\bf Inception & \bf \tabincell{c}{Inception-\\Text} & \bf \tabincell{c}{Deformable\\ PSROI pooling} & \bf Recall & \bf Precision & \bf F-measure \\ \hline
		\hline
		\xmark & \xmark & \xmark & 0.775 & 0.873 & 0.821 \\ \hline
		\cmark & \xmark & \xmark & 0.786 & 0.886 &0.833 \\ \hline
		\xmark & \cmark & \xmark & 0.803 & 0.891 & 0.845 \\ \hline
		
		\xmark & \cmark & \cmark & \bf 0.806 & \bf 0.905 & \bf 0.853 \\ \hline
		
	\end{tabular}
	\caption{Effectiveness of Inception-Text module and deformable PSROI pooling on ICDAR2015 incidental scene text location task.}
	\label{table:Impact}
\end{table}

\paragraph{Deformable PSROI pooling.} When we replace the standard PSROI pooling with deformable PSROI pooling, the precision has a large improvement (0.905 vs. 0.891). This indicates the power of our model for distinguishing text and non-text has been enhanced, more difficult regions have been correctly classified. After adding this module, the final F-measure increases from 0.845 to 0.853.

%In Figure \ref{fig:fmeasure}, we display the F-measure curves on ICDAR2015 with and without our proposed modules during training. According to the Figure, our proposed method is easier to train and has a better convergence point. 
%
%\begin{figure}[H]
%	\centering
%	\includegraphics[width=0.45\textwidth]{img/Fmeasure.jpg}
%	\caption{The F-measure curves on ICDAR2015 with and without Inception-Text module and Deformable PSROI pooling during training. The proposed method has a higher F-measure.}
%	\label{fig:fmeasure}
%\end{figure}

\subsubsection{Experiments on Scene Text Benchmarks.}

\begin{figure*}[th]
	\centering
	\subfigure[ICDAR 2015]{\includegraphics[height=0.3\textwidth]{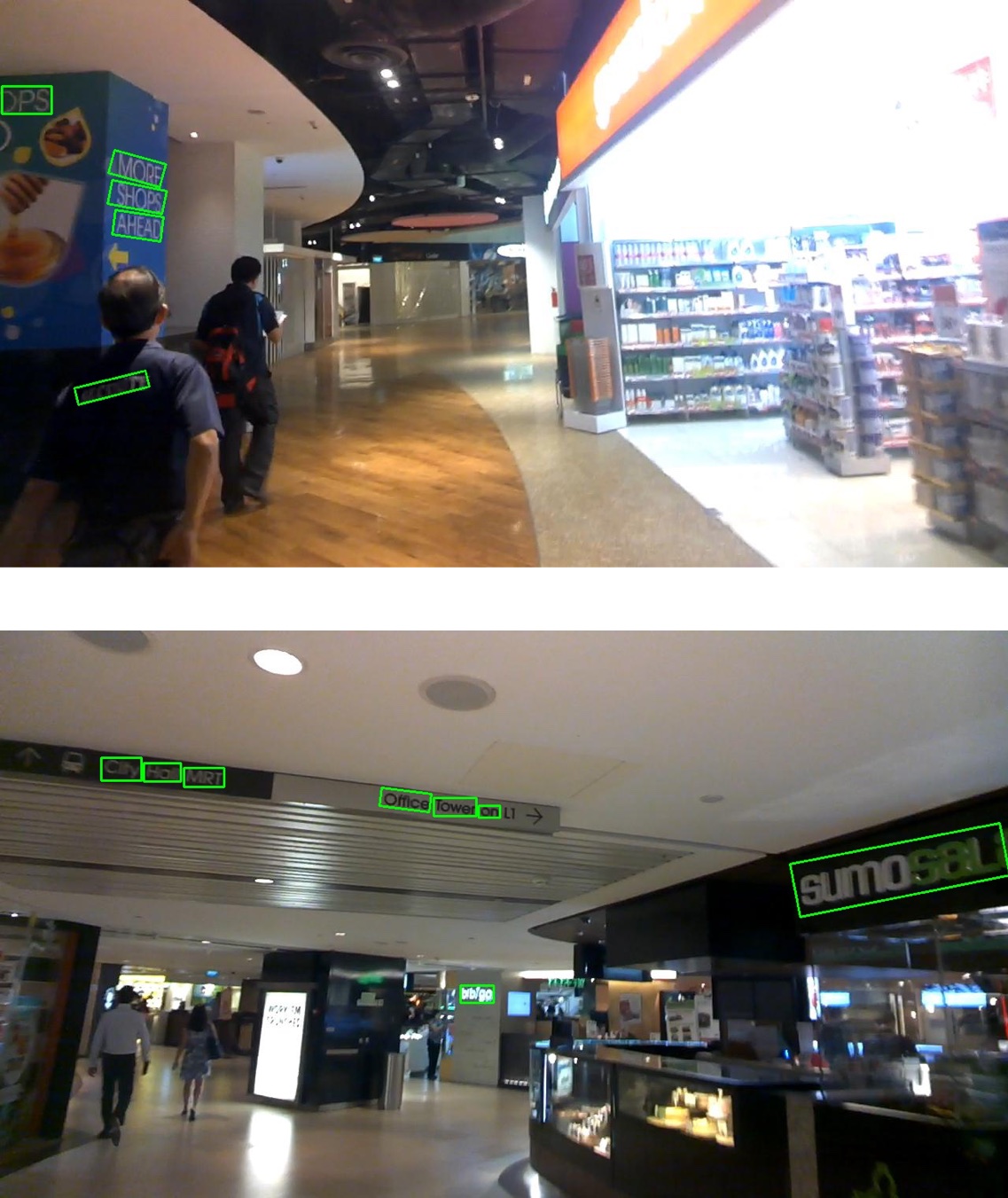}}
	\subfigure[RCTW-17]{\includegraphics[height=0.3\textwidth]{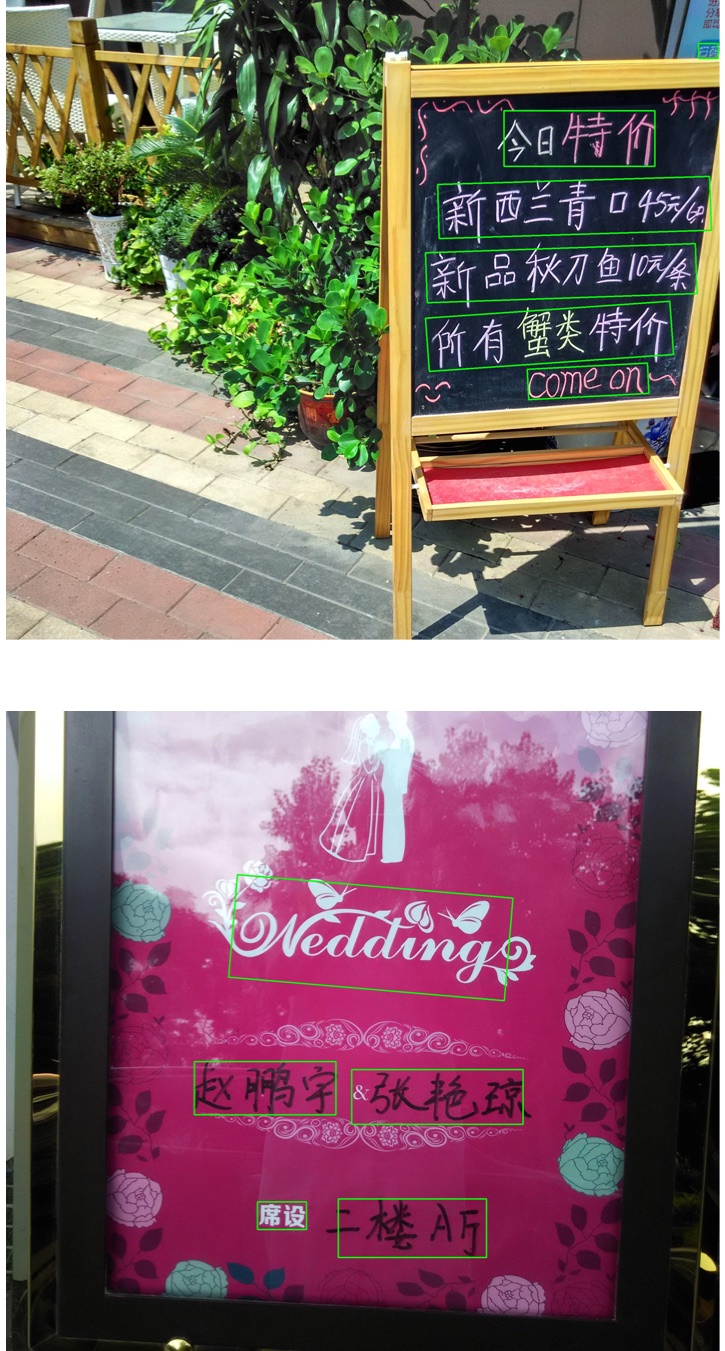}}
	\subfigure[MSRA-TD500]{\includegraphics[height=0.3\textwidth]{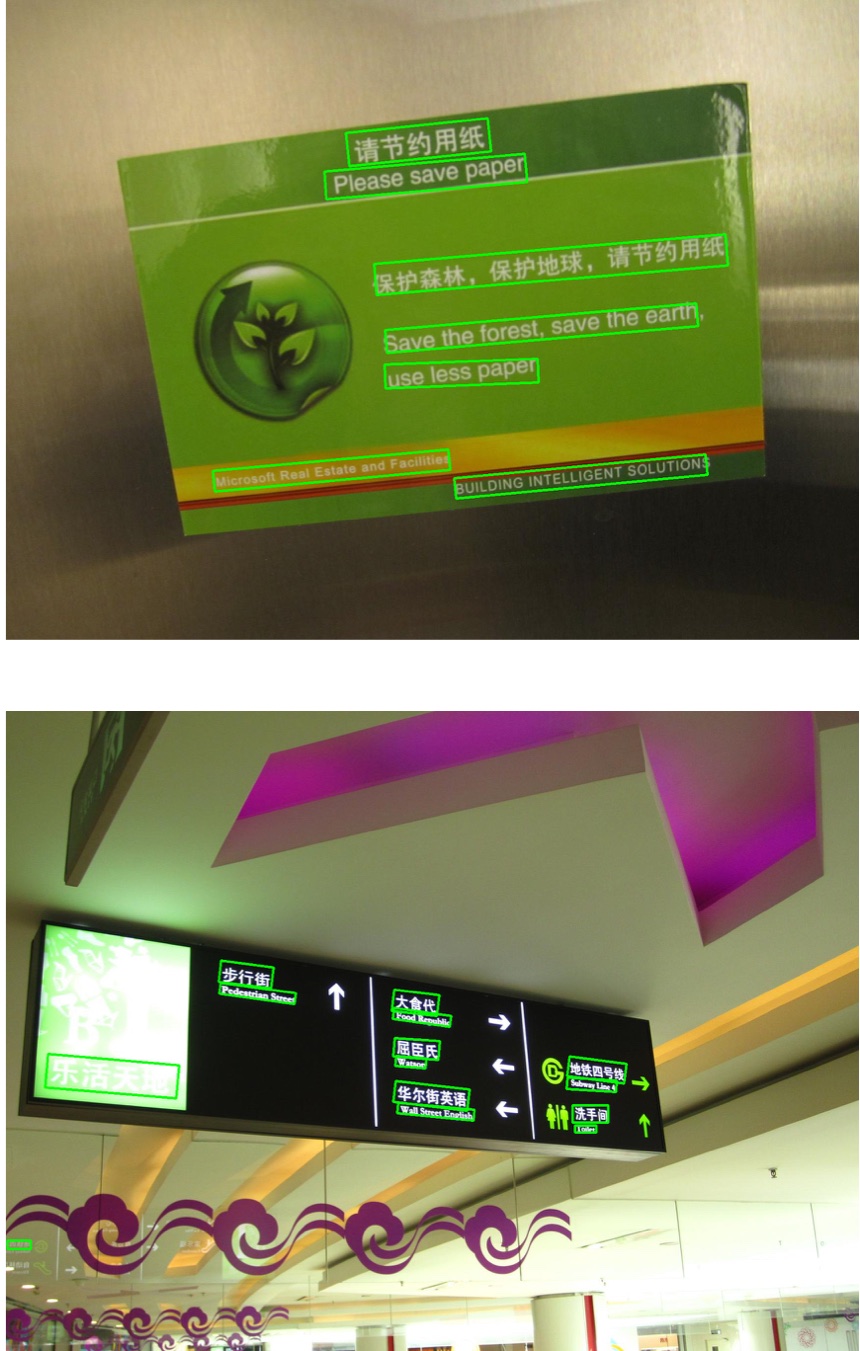}}
	\subfigure[Failure cases]{\includegraphics[height=0.3\textwidth]{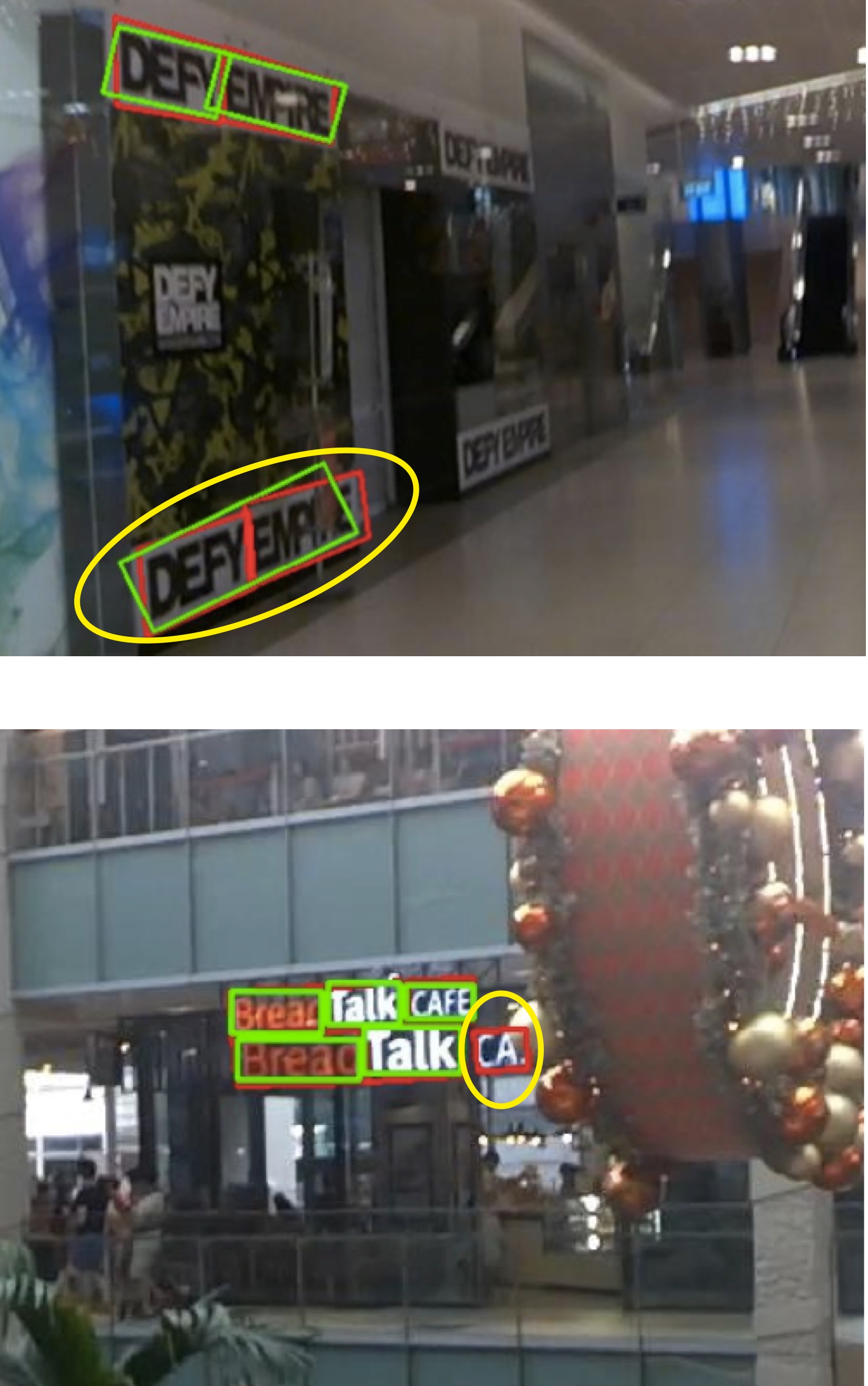}}
	\caption{Detection results of our proposed method on ICDAR2015(a), RCTW-17(b), MSRA-TD500(c). Some failure cases are presented in (d). Red boxes are ground-truth boxes, while green boxes ard the predict results. Bounding boxes in yellow ellipses represent the failures.}
	\label{fig:detect_results}
\end{figure*}

%We evaluate our proposed method on three real challenging scene text benchmarks: ICDAR2015, RCTW-17 and MSRA-TD500. These benchmarks all have scene text with arbitrary orientations.
We proceed to compare our method with the state-of-the-art methods on the public benchmark datasets, in Table~\ref{table:ICDAR2015} (ICDAR2015),  Table~\ref{table:RCTW17} (RCTW17) and Table~\ref{table:MSRATD500} (MSRA-TD500).
% Table \ref{table:ICDAR2015},  \ref{table:RCTW17} and \ref{table:MSRATD500} show the comparisons between our method and previous state-of-the-art methods.

On ICDAR2015, we only used $1000$ original images without extra data to train our network. With single scale of $960$, our proposed method achieves an F-measure of 0.853. When testing with two scales $[960, 1120]$, the F-measure is 0.868 which is over $0.02$ higher than the second best method in terms of absolute value. To the best of our knowledge, this is the best reported result in literature. 
Similar to~\cite{he2016deep}, we utilized an ensemble of 5 networks, while the backbones are ResNet101 (2 networks), ResNet50 (2 networks) and VGG (1 network). We used an ensemble of these 5 networks for proposing regions. And the union set of the proposals is processed by an ensemble for mask prediction and classification.
The final F-measure is 0.905, which is the 1st place result in ICDAR2015 leaderboard. \footnote{\url{http://rrc.cvc.uab.es/?ch=4&com=evaluation&task=1}}. 
The inference time of the ensemble approach is about 1.3s for the input image with resolution (1280 x 720) in ICDAR2015.

\begin{table}[th]
	\centering
	\fontsize{8}{8}\selectfont
	\setlength{\tabcolsep}{4pt}
	\renewcommand{\arraystretch}{2.0}
	\begin{tabular}{|l|c|r|r|r|}
		\hline
		\bf Method & \bf ExtraData & \bf Recall & \bf Precision & \bf F-measure \\ \hline \hline
		\bf IncepText ensemble & \xmark & \bf 0.873 & \bf 0.938 & \bf 0.905 \\ \hline
		\bf IncepText MS \footnotemark[3] & \xmark & 0.843 & 0.894 &  0.868 \\ \hline
		\bf IncepText & \xmark & 0.806 & 0.905 & 0.853 \\ \hline \hline
		FTSN \cite{dai2017fused} & \cmark & 0.800 & 0.886 & 0.841 \\ \hline
		R2CNN \cite{jiang2017r2cnn} & \cmark & 0.797 & 0.856 & 0.825 \\ \hline
		DDR \cite{he2017deep} & \cmark & 0.800 & 0.820 & 0.810 \\ \hline
		EAST \cite{zhou2017east} & \bf - & 0.783 & 0.832 & 0.807 \\ \hline
		RRPN \cite{ma2017arbitrary} & \cmark & 0.732 & 0.822 & 0.774 \\ \hline
		SegLink \cite{shi2017detecting} & \cmark & 0.731 & 0.768 & 0.749 \\ \hline
		\hline
	\end{tabular}
	\caption{Results on ICDAR2015 incidental scene text location task.}
	\label{table:ICDAR2015}
\end{table}

\begin{table}[th]
	\centering
	\fontsize{8}{8}\selectfont
	\setlength{\tabcolsep}{4pt}
	\renewcommand{\arraystretch}{2.0}
	\begin{tabular}{|l|c|r|r|r|}
		\hline
		\bf Method & \bf Recall & \bf Precision & \bf F-measure \\ \hline \hline
		\bf IncepText & \bf 0.569 & \bf 0.785 & \bf 0.660 \\ \hline
		FTSN \cite{dai2017fused} & 0.471 & 0.741 & 0.576 \\ \hline
		SegLink \cite{shi2017detecting} & 0.404 & 0.760 & 0.527 \\ \hline 
		\hline
	\end{tabular}
	\caption{Results on RCTW-17 text location task.}
	\label{table:RCTW17}
\end{table}

\begin{table}[th]
	\centering
	\fontsize{8}{8}\selectfont
	\setlength{\tabcolsep}{4pt}
	\renewcommand{\arraystretch}{2.0}
	\begin{tabular}{|l|c|r|r|r|}
		\hline
		\bf Method & \bf Recall & \bf Precision & \bf F-measure \\ \hline \hline
		\bf IncepText & \bf 0.790  & 0.875 & \bf 0.830 \\ \hline
		FTSN \cite{dai2017fused} & 0.771 & \bf 0.876 & 0.820 \\ \hline
		SegLink \cite{shi2017detecting} & 0.700 & 0.860 & 0.770 \\ \hline
		EAST \cite{zhou2017east} & 0.674 & 0.873 & 0.761 \\ \hline
		DDR \cite{he2017deep} & 0.700 & 0.770 & 0.740 \\ \hline
		%RRPN \cite{ma2017arbitrary} & 0.680 & 0.820 & 0.740 \\ \hline
		\hline
	\end{tabular}
	\caption{Results on MSRA-TD500 text location task.}
	\label{table:MSRATD500}
\end{table}

\footnotetext[3]{We only use two scales [960, 1120] of the short side.}

RCTW-17 is a new challenging benchmark on reading Chinese Text in images. Our proposed method achieves the F-measure of 0.66, which is a new state-of-the-art and significantly outperforms the previous methods.

On MSRA-TD500, our best performance achieves 0.790, 0.875 and 0.830 in recall, precision and F-measure, respectively. It exceeds the second best method by $0.01$ in terms of F-measure.

For the original resolution $(1280 \times 720)$ image in ICDAR2015, our proposed method takes about $270ms$ on a Nvidia Tesla M40 GPU.
The computation cost of the Inception-Text module is about $20ms$.

%\paragraph{Qualitative analysis.}

Some detection samples of our proposed method are visualized in Figure \ref{fig:detect_results}. In ICDAR2015, the text is mainly in word level, while the text is both in word and line level in RCTW-17 and MSRA-TD500. IncepText performs well in most situations, however it still fails in some difficult cases. A main limitations is that it fails to split two words with small word spacing, which is shown at the top of Figure~\ref{fig:detect_results} (d). Another weakness of IncepText is that it may miss the words which are occluded as illustrated at the bottom of Figure~\ref{fig:detect_results} (d).

\section{Conclusion}

In this paper, we proposed a novel end-to-end approach for multi-oriented scene text detection based on instance-aware segmentation framework. The main idea is to design a new Inception-Text module to handle scene text which suffers from a large variance of scale, aspect ratio and orientation. Another improvement comes from using deformable PSROI pooling to handle scene text. We demonstrated its efficiency on three public scene text benchmarks. Our proposed method achieves the state-of-the-art performance in comparison with the competing methods. As for future work, we would like to combine our detection framework with recognition framework to further boost the efficiency of our model.

%% The file named.bst is a bibliography style file for BibTeX 0.99c
\bibliographystyle{named}
\bibliography{inceptext}

\begin{thebibliography}{}

\bibitem[\protect\citeauthoryear{Abelson \bgroup \em et al.\egroup
  }{1985}]{abelson-et-al:scheme}
Harold Abelson, Gerald~Jay Sussman, and Julie Sussman.
\newblock {\em Structure and Interpretation of Computer Programs}.
\newblock MIT Press, Cambridge, Massachusetts, 1985.

\bibitem[\protect\citeauthoryear{Baumgartner \bgroup \em et al.\egroup
  }{2001}]{bgf:Lixto}
Robert Baumgartner, Georg Gottlob, and Sergio Flesca.
\newblock Visual information extraction with {Lixto}.
\newblock In {\em Proceedings of the 27th International Conference on Very
  Large Databases}, pages 119--128, Rome, Italy, September 2001. Morgan
  Kaufmann.

\bibitem[\protect\citeauthoryear{Brachman and
  Schmolze}{1985}]{brachman-schmolze:kl-one}
Ronald~J. Brachman and James~G. Schmolze.
\newblock An overview of the {KL-ONE} knowledge representation system.
\newblock {\em Cognitive Science}, 9(2):171--216, April--June 1985.

\bibitem[\protect\citeauthoryear{Gottlob \bgroup \em et al.\egroup
  }{2002}]{gls:hypertrees}
Georg Gottlob, Nicola Leone, and Francesco Scarcello.
\newblock Hypertree decompositions and tractable queries.
\newblock {\em Journal of Computer and System Sciences}, 64(3):579--627, May
  2002.

\bibitem[\protect\citeauthoryear{Gottlob}{1992}]{gottlob:nonmon}
Georg Gottlob.
\newblock Complexity results for nonmonotonic logics.
\newblock {\em Journal of Logic and Computation}, 2(3):397--425, June 1992.

\bibitem[\protect\citeauthoryear{Levesque}{1984a}]{levesque:functional-foundations}
Hector~J. Levesque.
\newblock Foundations of a functional approach to knowledge representation.
\newblock {\em Artificial Intelligence}, 23(2):155--212, July 1984.

\bibitem[\protect\citeauthoryear{Levesque}{1984b}]{levesque:belief}
Hector~J. Levesque.
\newblock A logic of implicit and explicit belief.
\newblock In {\em Proceedings of the Fourth National Conference on Artificial
  Intelligence}, pages 198--202, Austin, Texas, August 1984. American
  Association for Artificial Intelligence.

\bibitem[\protect\citeauthoryear{Nebel}{2000}]{nebel:jair-2000}
Bernhard Nebel.
\newblock On the compilability and expressive power of propositional planning
  formalisms.
\newblock {\em Journal of Artificial Intelligence Research}, 12:271--315, 2000.

\end{thebibliography}


\begin{thebibliography}{}

\bibitem[\protect\citeauthoryear{Busta \bgroup \em et al.\egroup
  }{2015}]{busta2015fastext}
Michal Busta, Lukas Neumann, and Jiri Matas.
\newblock Fastext: Efficient unconstrained scene text detector.
\newblock In {\em Proceedings of the IEEE International Conference on Computer
  Vision}, pages 1206--1214, 2015.

\bibitem[\protect\citeauthoryear{Dai \bgroup \em et al.\egroup
  }{2016}]{dai2016r}
Jifeng Dai, Yi~Li, Kaiming He, and Jian Sun.
\newblock R-fcn: Object detection via region-based fully convolutional
  networks.
\newblock In {\em Advances in neural information processing systems}, pages
  379--387, 2016.

\bibitem[\protect\citeauthoryear{Dai \bgroup \em et al.\egroup
  }{2017a}]{dai2017deformable}
Jifeng Dai, Haozhi Qi, Yuwen Xiong, Yi~Li, Guodong Zhang, Han Hu, and Yichen
  Wei.
\newblock Deformable convolutional networks.
\newblock {\em arXiv preprint arXiv:1703.06211}, 2017.

\bibitem[\protect\citeauthoryear{Dai \bgroup \em et al.\egroup
  }{2017b}]{dai2017fused}
Yuchen Dai, Zheng Huang, Yuting Gao, and Kai Chen.
\newblock Fused text segmentation networks for multi-oriented scene text
  detection.
\newblock {\em arXiv preprint arXiv:1709.03272}, 2017.

\bibitem[\protect\citeauthoryear{He \bgroup \em et al.\egroup
  }{2016a}]{he2016deep}
Kaiming He, Xiangyu Zhang, Shaoqing Ren, and Jian Sun.
\newblock Deep residual learning for image recognition.
\newblock In {\em Proceedings of the IEEE conference on computer vision and
  pattern recognition}, pages 770--778, 2016.

\bibitem[\protect\citeauthoryear{He \bgroup \em et al.\egroup
  }{2016b}]{he2016text}
Tong He, Weilin Huang, Yu~Qiao, and Jian Yao.
\newblock Text-attentional convolutional neural network for scene text
  detection.
\newblock {\em IEEE transactions on image processing}, 25(6):2529--2541, 2016.

\bibitem[\protect\citeauthoryear{He \bgroup \em et al.\egroup
  }{2017}]{he2017deep}
Wenhao He, Xu-Yao Zhang, Fei Yin, and Cheng-Lin Liu.
\newblock Deep direct regression for multi-oriented scene text detection.
\newblock {\em arXiv preprint arXiv:1703.08289}, 2017.

\bibitem[\protect\citeauthoryear{Jiang \bgroup \em et al.\egroup
  }{2017}]{jiang2017r2cnn}
Yingying Jiang, Xiangyu Zhu, Xiaobing Wang, Shuli Yang, Wei Li, Hua Wang, Pei
  Fu, and Zhenbo Luo.
\newblock R2cnn: Rotational region cnn for orientation robust scene text
  detection.
\newblock {\em arXiv preprint arXiv:1706.09579}, 2017.

\bibitem[\protect\citeauthoryear{Karatzas \bgroup \em et al.\egroup
  }{2015}]{karatzas2015icdar}
Dimosthenis Karatzas, Lluis Gomez-Bigorda, Anguelos Nicolaou, Suman Ghosh,
  Andrew Bagdanov, Masakazu Iwamura, Jiri Matas, Lukas Neumann,
  Vijay~Ramaseshan Chandrasekhar, Shijian Lu, et~al.
\newblock Icdar 2015 competition on robust reading.
\newblock In {\em Document Analysis and Recognition (ICDAR), 2015 13th
  International Conference on}, pages 1156--1160. IEEE, 2015.

\bibitem[\protect\citeauthoryear{Kingma and Ba}{2014}]{kingma2014adam}
Diederik Kingma and Jimmy Ba.
\newblock Adam: A method for stochastic optimization.
\newblock {\em arXiv preprint arXiv:1412.6980}, 2014.

\bibitem[\protect\citeauthoryear{Li \bgroup \em et al.\egroup
  }{2016}]{li2016fully}
Yi~Li, Haozhi Qi, Jifeng Dai, Xiangyang Ji, and Yichen Wei.
\newblock Fully convolutional instance-aware semantic segmentation.
\newblock {\em arXiv preprint arXiv:1611.07709}, 2016.

\bibitem[\protect\citeauthoryear{Long \bgroup \em et al.\egroup
  }{2015}]{long2015fully}
Jonathan Long, Evan Shelhamer, and Trevor Darrell.
\newblock Fully convolutional networks for semantic segmentation.
\newblock In {\em Proceedings of the IEEE Conference on Computer Vision and
  Pattern Recognition}, pages 3431--3440, 2015.

\bibitem[\protect\citeauthoryear{Ma \bgroup \em et al.\egroup
  }{2017}]{ma2017arbitrary}
Jianqi Ma, Weiyuan Shao, Hao Ye, Li~Wang, Hong Wang, Yingbin Zheng, and
  Xiangyang Xue.
\newblock Arbitrary-oriented scene text detection via rotation proposals.
\newblock {\em arXiv preprint arXiv:1703.01086}, 2017.

\bibitem[\protect\citeauthoryear{Neumann and Matas}{2010}]{neumann2010method}
Lukas Neumann and Jiri Matas.
\newblock A method for text localization and recognition in real-world images.
\newblock In {\em Asian Conference on Computer Vision}, pages 770--783.
  Springer, 2010.

\bibitem[\protect\citeauthoryear{Ren \bgroup \em et al.\egroup
  }{2015}]{ren2015faster}
Shaoqing Ren, Kaiming He, Ross Girshick, and Jian Sun.
\newblock Faster r-cnn: Towards real-time object detection with region proposal
  networks.
\newblock In {\em Advances in neural information processing systems}, pages
  91--99, 2015.

\bibitem[\protect\citeauthoryear{Shi \bgroup \em et al.\egroup
  }{2017}]{shi2017detecting}
Baoguang Shi, Xiang Bai, and Serge Belongie.
\newblock Detecting oriented text in natural images by linking segments.
\newblock {\em arXiv preprint arXiv:1703.06520}, 2017.

\bibitem[\protect\citeauthoryear{Szegedy \bgroup \em et al.\egroup
  }{2015}]{szegedy2015going}
Christian Szegedy, Wei Liu, Yangqing Jia, Pierre Sermanet, Scott Reed, Dragomir
  Anguelov, Dumitru Erhan, Vincent Vanhoucke, and Andrew Rabinovich.
\newblock Going deeper with convolutions.
\newblock In {\em Proceedings of the IEEE conference on computer vision and
  pattern recognition}, pages 1--9, 2015.

\bibitem[\protect\citeauthoryear{Szegedy \bgroup \em et al.\egroup
  }{2016}]{szegedy2016rethinking}
Christian Szegedy, Vincent Vanhoucke, Sergey Ioffe, Jon Shlens, and Zbigniew
  Wojna.
\newblock Rethinking the inception architecture for computer vision.
\newblock In {\em Proceedings of the IEEE Conference on Computer Vision and
  Pattern Recognition}, pages 2818--2826, 2016.

\bibitem[\protect\citeauthoryear{Szegedy \bgroup \em et al.\egroup
  }{2017}]{szegedy2017inception}
Christian Szegedy, Sergey Ioffe, Vincent Vanhoucke, and Alexander~A Alemi.
\newblock Inception-v4, inception-resnet and the impact of residual connections
  on learning.
\newblock In {\em AAAI}, pages 4278--4284, 2017.

\bibitem[\protect\citeauthoryear{Tian \bgroup \em et al.\egroup
  }{2016}]{tian2016detecting}
Zhi Tian, Weilin Huang, Tong He, Pan He, and Yu~Qiao.
\newblock Detecting text in natural image with connectionist text proposal
  network.
\newblock In {\em European Conference on Computer Vision}, pages 56--72.
  Springer, 2016.

\bibitem[\protect\citeauthoryear{Xie and Tu}{2015}]{xie2015holistically}
Saining Xie and Zhuowen Tu.
\newblock Holistically-nested edge detection.
\newblock In {\em Proceedings of the IEEE international conference on computer
  vision}, pages 1395--1403, 2015.

\bibitem[\protect\citeauthoryear{Yao \bgroup \em et al.\egroup
  }{2012}]{yao2012detecting}
Cong Yao, Xiang Bai, Wenyu Liu, Yi~Ma, and Zhuowen Tu.
\newblock Detecting texts of arbitrary orientations in natural images.
\newblock In {\em Computer Vision and Pattern Recognition (CVPR), 2012 IEEE
  Conference on}, pages 1083--1090. IEEE, 2012.

\bibitem[\protect\citeauthoryear{Yao \bgroup \em et al.\egroup
  }{2016}]{yao2016scene}
Cong Yao, Xiang Bai, Nong Sang, Xinyu Zhou, Shuchang Zhou, and Zhimin Cao.
\newblock Scene text detection via holistic, multi-channel prediction.
\newblock {\em arXiv preprint arXiv:1606.09002}, 2016.

\bibitem[\protect\citeauthoryear{Zhou \bgroup \em et al.\egroup
  }{2017}]{zhou2017east}
Xinyu Zhou, Cong Yao, He~Wen, Yuzhi Wang, Shuchang Zhou, Weiran He, and Jiajun
  Liang.
\newblock East: An efficient and accurate scene text detector.
\newblock {\em arXiv preprint arXiv:1704.03155}, 2017.

\end{thebibliography}

\end{document}